\documentclass[letterpaper, 10 pt, conference]{ieeeconf}  

\IEEEoverridecommandlockouts                              

\usepackage{hyperref}
\hypersetup{hidelinks,
	colorlinks=true,
	allcolors=black,
	pdfstartview=Fit,
	breaklinks=true}
\usepackage{graphics} 
\usepackage{epsfig} 
\usepackage{mathptmx} 
\usepackage{times} 
\usepackage{amsmath} 
\usepackage{amssymb}  

\title{\LARGE \bf
Unsupervised OmniMVS: Efficient Omnidirectional Depth Inference via Establishing Pseudo-Stereo Supervision
}

\author{Zisong Chen, Chunyu Lin$^{\dagger}$, Lang Nie, Kang Liao, and Yao Zhao
\thanks{Institute of Information Science, Beijing Jiaotong University, Beijing 100044, China. Email: \{zschen, cylin,nielang, kang\_liao,  yzhao\}@bjtu.edu.cn
}
\thanks{$^{\dagger}$ Corresponding author.
}
}
\usepackage{graphicx}
\usepackage{amsmath}
\usepackage{ulem}
\begin{document}
\maketitle
\thispagestyle{empty}
\pagestyle{empty}
\begin{abstract}
Omnidirectional multi-view stereo (MVS) vision is attractive for its ultra-wide field-of-view (FoV), enabling machines to perceive 360° 3D surroundings. However, the existing solutions require expensive dense depth labels for supervision, making them impractical in real-world applications.

In this paper, we propose the first unsupervised omnidirectional MVS framework based on multiple fisheye images. 
To this end, we project all images to a virtual view center and composite two panoramic images with spherical geometry from two pairs of back-to-back fisheye images. The two 360° images formulate a stereo pair with a special pose, and the photometric consistency is leveraged to establish the unsupervised constraint, which we term ``Pseudo-Stereo Supervision". 
In addition, we propose Un-OmniMVS, an efficient unsupervised omnidirectional MVS network, to facilitate the inference speed with two efficient components.
First, a novel feature extractor with frequency attention is proposed to simultaneously capture the non-local Fourier features and local spatial features, explicitly facilitating the feature representation. Then, a variance-based light cost volume is put forward to reduce the computational complexity. Experiments exhibit that the performance of our unsupervised solution is competitive to that of the state-of-the-art (SoTA) supervised methods with better generalization in real-world data. The code will be available at \url{https://github.com/Chen-z-s/Un-OmniMVS}.
\end{abstract}
\section{INTRODUCTION}
\label{s1}

Obtaining depth about the surroundings is an important task to perceive the real 3D environment, which is widely demanded in indoor robotic systems \cite{r41}, autonomous driving \cite{r42},  virtual reality \cite{r43}, etc. There are various depth acquisition devices, including LiDAR, structured light 3D scanners, stereo cameras, etc. Among these vision systems, camera-based configurations are preferred by researchers because they are compact, convenient, inexpensive, and can provide dense depth maps. Besides, with the development of deep learning, it has become increasingly popular to estimate the depth directly from camera images. However, most researches focus on predicting depth from normal FoV cameras \cite{r22,r23,r24}, failing to perceive the 360° depth, which might result in wrong decisions to avoid obstacles or plan trajectories in a robotic vision system.  

\begin{figure}
\centering
\includegraphics[scale=0.59]{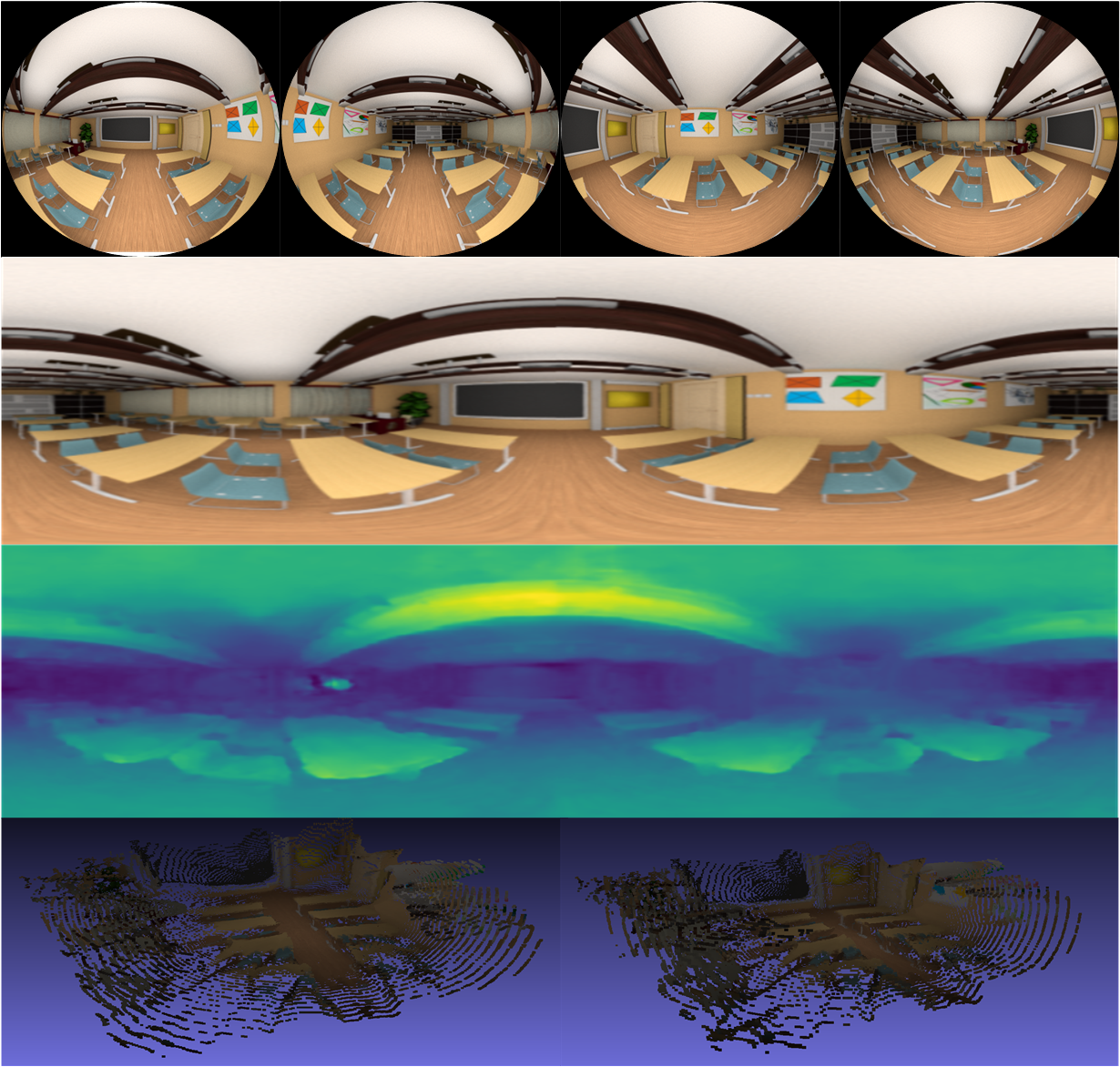}
\caption{\label{Fig1}
Row 1: Four fisheye inputs. Row 2: Projected panorama using the predicted depth map. Row 3: Inverse depth map. Row 4: Point cloud.}
\end{figure}

Recently, a new but practical omnidirectional MVS configuration is proposed in \cite{r10,r17}, where four fisheye lenses with a 220° FoV are placed to face four cardinal directions. 
To generate omnidirectional depth maps with a 360° horizontal FoV and a 180° vertical FoV (ERP format), a spherical scanning strategy based on the device center is proposed, which can produce seamless omnidirectional depth maps without manual division of views and subsequent stitching. However, all these methods require dense depth labels to supervise the training process of the neural network. To obtain the ground truth, expensive acquisition devices such as LiDAR and extensive manual labor are required, making these supervised solutions impractical.

To solve the above problem, we propose the first unsupervised omnidirectional MVS framework based on multiple fisheye images. 
Motivated by the unsupervised constraint in traditional stereo matching, we propose to establish pseudo-stereo supervision for our distinct camera setup. 
First, we define the device center as a new virtual view center, and project two pairs of back-to-back fisheye images (Row 1, Fig. \ref{Fig1}) to this view according to the spherical geometry with predicted depth. Then, these projected images can be composited as two panoramic images (Row 2, Fig. \ref{Fig1}). The two synthetic panoramas formulate a stereo pair with a special camera pose, where the baseline is 0. In this case, the two synthetic views can be aligned by pure rotation. Therefore, we leverage the photometric consistency between one synthetic panorama and the other rotated one to constrain the predicted depth (Row 3, Fig. \ref{Fig1}), establishing pseudo-stereo supervision without dense depth labels. 
\par

Furthermore, we design an efficient unsupervised omnidirectional MVS network, named Un-OmniMV. We follow the pipeline of OmniMVS \cite{r17} and then propose an efficient feature extractor with a frequency attention module.
Particularly, we employ convolution
operations in the frequency domain to highlight frequency domain components of interest (non-local features) and fuse them with local features extracted from the spatial domain. 
Compared with the conventional backbones \cite{r34,r49}, our solution adopts fewer downsampling operations but ensures sufficient receptive fields, revealing both effectiveness and efficiency.
In addition, to reduce the computational complexity in the cost volume establishment, we design a variance-based light cost volume to squeeze the traditional cost volume \cite{r17} to a quarter with variance representation.
Extensive experiments demonstrate our superiority over other SoTA solutions. The contributions can be summarized as follows:
\begin{itemize}
    \item 
    To free the demand for expensive depth labels, we propose pseudo-stereo supervision, contributing to the first unsupervised omnidirectional MVS solution.
    \item We propose Un-OmniMVS, an efficient unsupervised omnidirectional MVS network, to facilitate the inference speed via frequency attention module and variance-based light cost volume.
    \item The proposed unsupervised solution achieves competitive performance to the SoTA supervised schemes with better generalization, especially in real-world data. 
\end{itemize}
\begin{figure}
\centering
\includegraphics[scale=0.7]{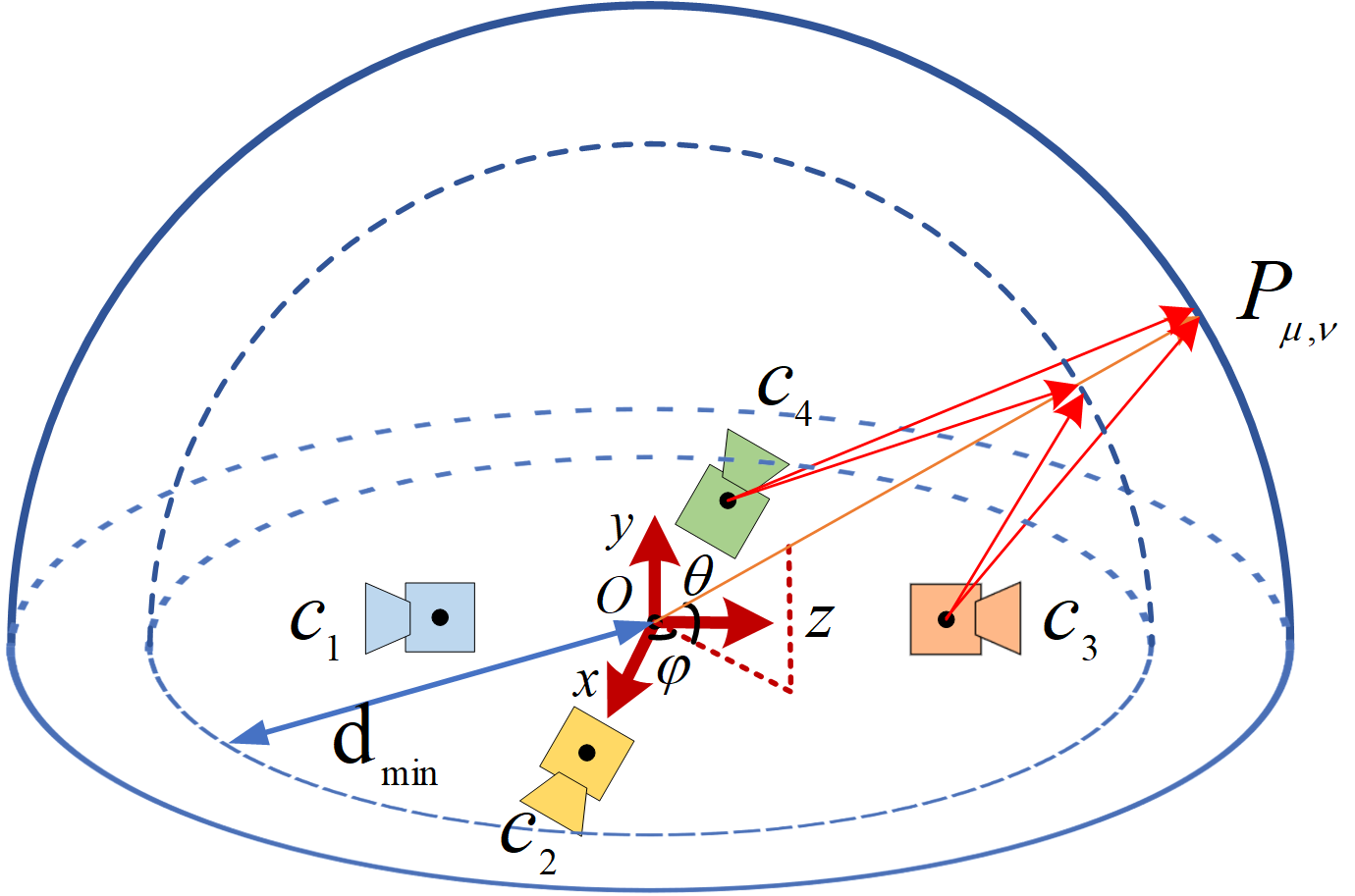}
\caption{\label{Fig2}\textbf{Device configuration:} The wide baseline multi-camera rig system is equipped with four fisheye lenses with 220° FoV. The spherical scanning volume corresponding to the fisheye image can be generated by projection according to different depths. Only part of the sphere is shown here.}
\end{figure}
\begin{figure*}[ht]
\centering
\includegraphics[scale=0.45]{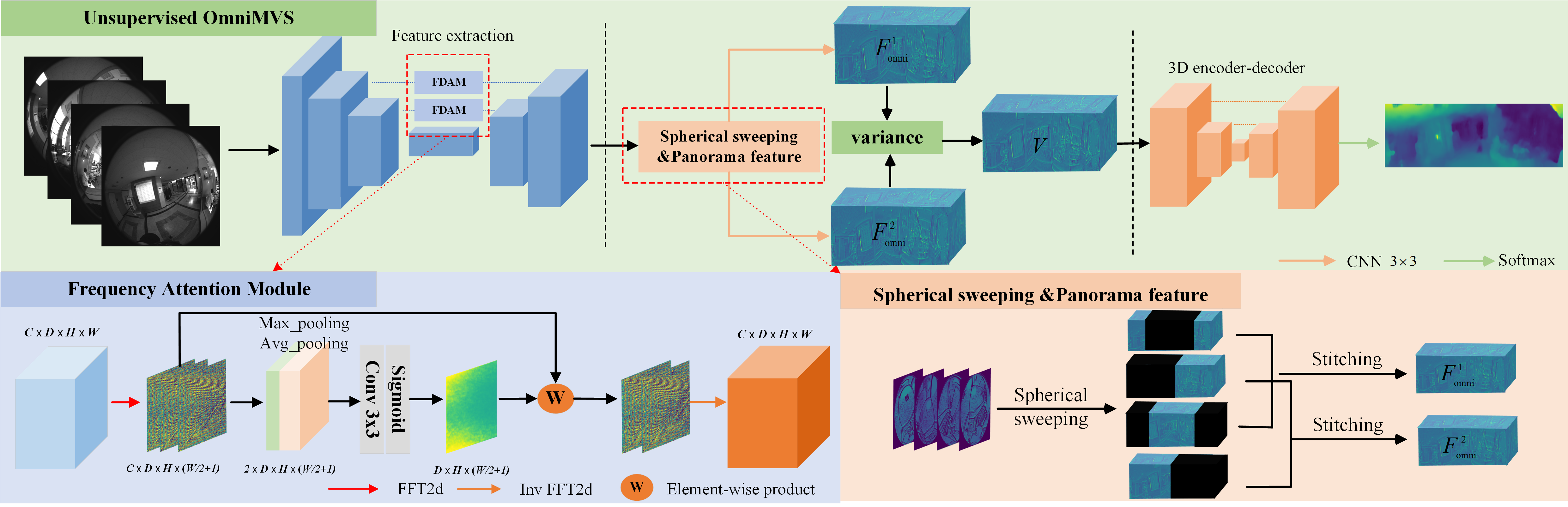}
\caption{\label{Fig3}The architecture of our proposed Unsupervised OmniMVS. It contains three components: Frequent-Spatial Feature Extractor, Variance-based cost volume generation, and 3D encoder and decoder. Frequency Attention Module (FAM): In the frequency domain, the network captures the interested frequency domain components based on the attention mechanism. Spherical sweeping \& Panorama Feature: Complete the projection of input features to spherical features and the stitching of panoramic features.}
\label{fig_framework}
\end{figure*}

\section{Related work}
\subsection{Multi-Fisheye Omnidirectional Depth Estimation}
The current supervised omnidirectional depth estimation algorithms are all based on deep learning. In many works, the environment is directly captured by omnidirectional cameras \cite{r4,r5,r7,r8}, and the neural network is used to implicitly learn the structural information of panoramic images to predict depth information.
Some panorama work also uses multiple cameras \cite{r10,r11,r17,r21,r52,r53} to realize the depth perception of the panorama as shown in Fig. \ref{Fig2}. Meuleman et al. \cite{r52} design an adaptive spherical matching method and propose a fast inter-scale bilateral cost volume filtering method, which enables real-time dense distance estimation without relying on depth information. SweepNet \cite{r10} and OmniMVS \cite{r17} proposed a novel wide-baseline omnidirectional stereo vision system that estimates omnidirectional and continuous depth maps using only 4 cameras. SweepNet \cite{r10} proposes to calculate the matching cost between multiple spherical features by SGM. OmniMVS \cite{r17} combines deep learning methods to propose an end-to-end deep neural network, which uses 3D encoding and decoding to achieve regularization of the matching cost and then obtains the depth map through softmax regression. Based on previous work, Chang $et\ al.$ \cite{r21} introduced an uncertainty regularization term to alleviate the edge blurring of predictions. Considering that only sparse depth information can be obtained through radar in practice, Lee et al. \cite{r53} propose a semi-supervised learning method that combines sparse depth supervision information and pixel matching on overlapping views, which alleviates the reduction in prediction performance caused by sparse supervision information.\par
\subsection{Stereo Unsupervised Depth Estimation}
Under the narrow field of the view camera system, one form of self-supervision comes from the stereo composition. During training, the deep network \cite{r12,r13,r14,r15} is trained based on the pixel difference between the original image and the synchronously reprojected image pair according to the geometric relationship, and monocular depth estimation can be achieved at test time. Based on a wide-field camera, a stereo-matching model for binocular fisheye is proposed \cite{r40}.
To obtain a full range of depth information, Zioulis et al. \cite{r16} proposed to use panoramic image pairs, combined with image parallax and spherical geometric relationships to calculate the disparity between images. Unsupervised training is achieved through a geometric depth-driven photometric image reconstruction loss and a depth smoothing loss, similar to how ordinary camera stereo pairs are matched.\par
Despite that, the unsupervised learning of omnidirectional depth inference with multiple fisheye cameras \cite{r17} has never been investigated due to the special camera setup.
\section{Unsupervised Spherical Depth}
\label{section3}

\subsection{Spherical Geometry}
\label{section20}
As illustrated in Fig. \ref{Fig2}, a world coordinate system is established in our device configuration with the rig center as the origin, where the $x$-$z$ plane is aligned with the camera center.
Considering the goal to obtain the panoramic depth, we represent the omnidirectional depth estimation system in a sphere and leverage the longitude ($\varphi$) and latitude ($\theta$) to describe that, where $\varphi\in(-180^\circ,180^\circ)$ and $\theta\in(-90^\circ,90^\circ)$.
Then an arbitrary point in the spherical space can be defined as the following equation with $\theta$, $\varphi$ and $d$: 
\begin{equation}
P = d[cos(\varphi)cos(\theta),\ sin(\varphi),\ cos(\varphi)sin(\theta)]^T,
\end{equation}
where $d$ is the radius of the sphere and it is also the distance from the sphere point to the origin of the coordinates.

$C_i$ is the $i$-th camera in our system with the rotation $R_i$ and translation $T_i$ to the rig center. The point $P_i$ in the camera coordinate system can be obtained from the point $P$ in the world coordinate system through the following formulation:
\begin{equation}
P_i = R_iP+T_i\label{eq2}.
\end{equation}
Given the intrinsic parameters of the camera, the projection relationship between the spatial point $ P_i$ and the corresponding pixel position $p_i$ on the fisheye image can be established through a projection function $\prod_{i}(\cdot)$.
For an arbitrary point $P$, we can find the corresponding pixel position $p_i$ in the $i$-th fisheye image as Eq. (\ref{eq4}):
\begin{equation}
p_i=\prod_{i}(R_iP+T_i)\label{eq4}.
\end{equation}

According to the above equation, the value of $P$ can be obtained by applying differentiable bilinear interpolation to the fisheye image as follows:
\begin{equation}
V_P=G(I_i, \prod_{i}(R_iP+T_i))\label{eq5},
\end{equation}
where $G(\cdot,\cdot)$ denotes the interpolation operation.

In our spherical system, the sphere can be easily mapped into an equirectangular image $S^{d}$. Assuming the resolution of this image is $H \times W$, the sampling intervals of latitude and longitude are $\delta_1=180^\circ/H$ and $\delta_2=360^\circ/W$. Given the pixel coordinate $(\mu,\nu)$ in this image, we can map it to a 3D point in spherical space $P^{d}_{\mu,\nu}=d[cos(\mu\delta_1)cos(\nu\delta_2),sin(\mu\delta_1),cos(\mu\delta_1)sin(\nu\delta_2)]^T$, and get the projected pixel value.
In this manner, the fisheye image of the $i$-th camera can be warped to the equirectangular image through a sphere with the specific depth $d$ as:
\begin{equation}
S^{d}_i=\bigcup_{\mu,\nu=0}^{H,W} G(I_i, \prod_{i}(R_iP^{d}_{\mu,\nu}+T_i)) \label{eq5}.
\end{equation}
\begin{figure}[ht]\centering
\includegraphics[scale=0.35]{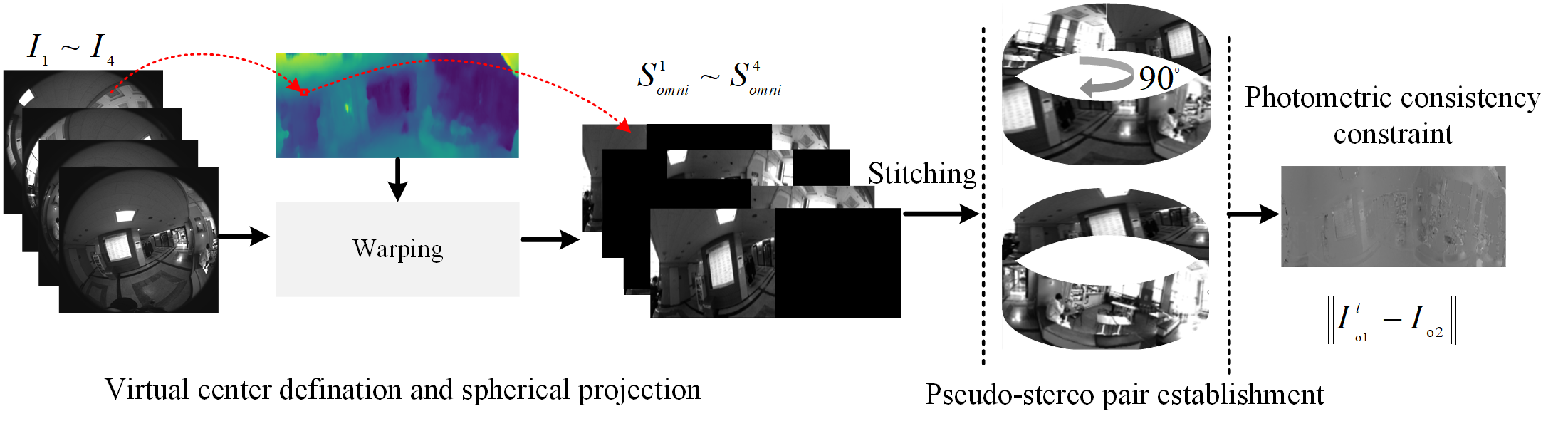}
\caption{ \label{Fig4} The proposed "Pseudo-Stereo Supervision" is based on spherical stereo matching. Through end-to-end training, omnidirectional disparity (inverse depth) prediction is obtained, and an omnidirectional stereo pair with a special pose is established by spherical back-projection and panoramic composition.}
\end{figure}
\subsection{Network Structure}
\label{section20}
In this work, we design the feature extraction, feature matching, and depth regression stages as our basic pipeline.
Particularly, we introduce frequency domain convolution in the feature extraction and a variance-based light cost volume proposed to represent the feature correlation with decreased spatial complexity. Finally, we leverage the 3D codec layer and softmax \cite{r20} regression to predict the disparity map $D(\theta,\varphi)$.\par
\subsubsection{Frequent-Spatial Feature Extractor}
\label{section30}
ResNet \cite{r46} and UNet \cite{r45} usually serve as backbones \cite{r34,r49} for various vision tasks. To enhance the receptive fields of the networks, extensive downsampling operations are adopted. But in fisheye images, fisheye distortions would severely squeeze abundant contents into small regions. So, we argue these popular backbones (including excessive downsampling operations) would lose the detailed information, thus leading to inferior performance. Although some unconventional convolutions \cite{r51} can be introduced to facilitate the perception of distorted contents, it brings a large computational burden.

To solve the above issues, we introduce the frequency attention module to achieve global information acquisition. The detailed structure is illustrated in Fig. \ref{Fig3}, where the spatial domain features are first transformed into the frequency domain through  FFT. Avg-pooling and max-pooling are used to squeeze the spatial dimension of frequency domain features, followed by a concatenation operation. Then, a convolution with a sigmoid function is used to learn the frequency attention map. After the element-wise product, we extract the frequency domain components of interest. Considering the local frequency domain features can cover the receptive fields of the entire spatial domain features, the frequency domain convolution effectively captures the non-local information without downsampling. After that, an inverse FFT is performed to restore the spatial domain feature. With the frequency attention module, our feature extractor is efficient with only two downsampling operations.
\subsubsection{Light Cost Volume}
\label{section30}
According to Eq. (\ref{eq5}), we can obtain four panoramic feature maps with the size of $C\times N\times H^f\times W^f$ by projecting four fisheye features to N depth planes, as shown in Fig. \ref{Fig3}. In OmniMVS \cite{r17}, four spherical feature volumes are directly stacked to form a $4C\times N\times H^f\times W^f$ cost volume. However, conducting 3D convolutions on such a large cost volume requires lots of GPU memory and computational resources. To alleviate it, Won et al. \cite{r21} crop each view to 180° after projection, and these spherical features are interleaved and concatenated into a $2C\times N\times H^f\times W^f$ cost volume, which cuts computational complexity in half. \par


 Motivated by multi-view reconstruction \cite{r22}, we propose a more lightweight cost volume based on variance, as shown in Fig. \ref{Fig3}. First, we establish the four spherical feature volumes from the fisheye features and crop their FoV to 180°. Then, two 360° spherical feature volumes can be obtained by stitching the four 180° feature volumes. Next, we do not directly calculate the variance between the spherical feature volumes. Instead, a $3 \times 3$ convolution is carried out to gather the $3 \times 3$ neighborhood features to one point before calculating the variance. The two processed panoramic feature volumes are more discriminative and can be denoted as $F_{omni}^{1}$, $F_{omni}^{2}$. Our cost volume is calculated as the following formulation:
 
\begin{equation}
V =\frac{\sum_{i=1}^{2}(F_{omni}^{i}-\overline{F})^2}{2},
\end{equation}
where $\overline{F}$ is the average volume of $F_{omni}^{1}$ and $F_{omni}^{2}$.
We end up with a cost volume of size $C\times N\times H^f\times W^f$, which is a quarter of that in OmniMVS.
\subsubsection{Disparity regression}
In the last stage, we leverage the 3D codec to process the cost volume and get a single-channel volume $V^*$ with the size of $N\times H\times W$. Then the disparity map can be obtained by softargmin as \cite{r44}:
\begin{equation}
D(\theta,\varphi)=\sum_{D=D_{min}}^{D_{max}}D\times softmax(V^*(\theta,\varphi,D)),
\end{equation}
where $D_{max}$ and $D_{min}$ are the maximum and minimum values of the disparity. Depth map is $D_{depth}=1/D(\theta,\varphi)$.
\subsection{Pseudo-Stereo Supervision}
\label{section20}
\subsubsection{Virtual Center and Projection} To establish pseudo-stereo supervision, we first define the device center as the common virtual view. The four fisheye images can be projected into this view as equirectangular images $S_{omni}^{i} (i=1,2,3,4)$ with the predicted depth map $D_{depth}$, as shown in Fig. \ref{Fig4}. 
Specifically, given a panoramic image with the resolution $H \times W$ and the depth value $d_{\mu,\nu}$ of each pixel from $D_{depth}$, the 3D spatial position can be represented as $P_{\mu,\nu}^{d_{\mu,\nu}}$. Then we can calculate the corresponding pixel position in the $i$-th fisheye image by Eq. (\ref{eq4}). The fisheye image can be projected onto the corresponding equirectangular image $S_{omni}^{i}$ through Eq. (\ref{eq5}) with $d$ being changed to predicted depth $d_{\mu,\nu}$.

\subsubsection{Pseudo-Stereo Pair} We crop $S_{omni}^{i}$ to limit its FoV to 180°. Then we stitch the four equirectangular images with 180° FoV to form two panoramas with 360° FoV. Concretely, every panoramic image is composited of two projected images from back-to-back cameras as follows:
\begin{equation}
I_{o1},I_{o2}=S(S_{omni}^{i},S_{omni}^{i+2}) \ i\in(1,2),\label{eq8}
\end{equation}
where $S(\cdot)$ is the stitching operation.
$I_{o1}, I_{o2}$ are two stitched panoramic images. Now, we formulate a pseudo-stereo pair with a special camera pose. In this case, the baseline between virtual cameras is 0$cm$, and the yaw is 90°.

\subsubsection{Spherical Pseudo-Stereo Matching} To match the two 360° images, we need to rotate one ($I_{o1}$) 90° around the yaw-axis on the sphere. Actually, the rotation around the yaw-axis on the sphere can be simplified as the circular translation in the ERP format. Ideally, the translated image ($I^{t}_{o1}$) could align with the other ($I_{o2}$) perfectly if the predicted depth is absolutely correct. 
Besides, the proposed pseudo-stereo constraint enables the network to directly predict the actual spatial depth while avoiding the scale uncertainty caused by supervised methods.

\subsection{Unsupervised Optimization: }
\label{section30}
The total optimization goal is defined as follows ($\beta_1=1$, $\beta_2=2$, $\beta_3=1$.):
\begin{equation}
L_{total}=\beta_1 L_p + \beta_2 L_s + \beta_3 L_{g},\label{eq13},
\end{equation}
where $L_p$, $L_s$ and $L_{g}$ are the photometric loss, smoothness loss, and gradient loss. 

\subsubsection{Photometric Loss}
\label{section30}
With the proposed pseudo-stereo supervision, we can establish the unsupervised constraint by photometric consistency. 
To encourage the similarity at local pixel and global structure simultaneously, we combine the L1 and SSIM \cite{r24} to formulate the photometric loss $L_{p}$:
\begin{equation}
L_{p}=\frac{\alpha}{2}(1-SSIM(I^{t}_{o1},I_{o2}))+(1-\alpha)\parallel I^{t}_{o1}-I_{o2}\parallel,
\end{equation}
where $\alpha$ is a hyperparameter and empirically set to 0.85.
\subsubsection{Edge-Aware Smoothness Loss}
\label{section30}
The values in the predicted disparity map around the textureless and occluded areas tend to be divergent. To smooth this uncertain scatter, Eq. (\ref{eq10}) is usually used to smooth the depth gradient \cite{r12,r15} with the input image as reference:
\begin{equation}
L_{s}(d_o,I_o)=|\partial_{x}d_o|e^{-|\partial_{x}I_o|}+|\partial_{y}d_o|e^{-|\partial_{y}I_o|},\label{eq10}
\end{equation}
where $d_o$ and $I_o$ denote the disparity map and input image.
Therefore, we can get the smoothness loss with two panorama images $I^{t}_{o1},I_{o2}$ and disparity map $D(\theta,\varphi)$:
\begin{equation}
L_{s}=L_{s}(D(\theta,\varphi),I^{t}_{o1})+L_{s}(D(\theta,\varphi),I_{o2}).
\end{equation}
\subsubsection{Gradient Loss}
\label{section30}
Shooting from different perspectives usually results in noticeable chromatic aberration, which brings a new challenge to the proposed unsupervised photometric constraint. Therefore, we employ the gradients of the two reconstructed panoramas for further supervision.
\begin{equation}
L_{g}=(\partial_{x}I^{t}_{o1}-\partial_{x}I_{o2})+(\partial_{y}I^{t}_{o1}-\partial_{y}I_{o2}).
\end{equation}
\def\tablename{\large Table}
\section{Experiments}
\label{section4}
\subsection{Dataset and Metrics:}
\label{section30}
\subsubsection{Dataset}
OmniMVS \cite{r17} provides three synthetic datasets, in which depth labels, extrinsic parameters, intrinsic parameters, and four fisheye images ($H_{1}=768, W_{1}=800$) are provided. Here we briefly introduce these datasets used in our experiments.

\begin{itemize}
    \item The OmniThings contains 10,240 diverse scenes, and it is the largest and richest dataset. We select 4000 images as the training set and 1000 images as the test set.
    \item The OmniHouse generates realistic interior scenes, including 2560 house models. We select 2000 images as the training set and 500 images as the test set.
    \item The Sunny provides 1000 outdoor street scene data. We select 800 images as the training set and 200 images as the test set.
\end{itemize}

\subsubsection{Metrics}\label{section30}
To quantitatively compare our method with previous spherical sweeping methods \cite{r27,r17,r47}, we adopt the popular evaluation metrics in previous works \cite{r17} as follows:

\begin{equation}
E(\theta,\varphi)=\frac{\vert D^{*}_{gt}(\theta,\varphi) - D^{*}(\theta,\varphi) \vert}{N}.
\end{equation}

More specifically, we employ MAE, MSE, and error ratios$(\%)$ larger than $n (\textgreater n)$ for quantitative evaluation, where $n$ is 1, 3, and 5 respectively.\par
\subsection{Implement Details and Inference Speed}
\label{section30}
We use PyTorch to implement the proposed end-to-end network. During training, the number of spheres N is set to 32, and the number of cost volume channels C is set to 32. The minimum distance is set to 0.55 meters, and the maximum distance is set to $10^{5}$ meters. The spherical depth is selected according to the 32 equal divisions of the inverse depth and the disparity map resolution is set to $640 \times320$. Our network is trained using an Adam optimizer \cite{r39} with an exponentially decaying learning rate initialized to $1 \times 10^{-4}$ for the 30 epochs. 
In the inference, it takes 0.33s for our method to accomplish an iteration, which is 2x faster than the 0.67s of OmniMVS.
\subsection{Experimental Results}
\subsubsection{Comparison Experiment}
We compare the proposed method with ZNCC + SGM \cite{r27}, MC-CNN \cite{r47}+ SGM \cite{r27}, and OmniMVS \cite{r17}. Among these methods, ZNCC and MC-CNN use patches with $9 \times 9$ to calculate the matching cost of different spherical images, and SGM regularizes the cost volume by adding smoothness constraints. OmniMVS uses a neural network to calculate the matching cost in the overall spherical feature directly. The quantitative comparisons are shown in Table \ref{lable1}, where $S$ and $U$ denote ``supervised" and ``unsupervised". From this table, we can observe:
\begin{itemize}
    \item Our method is better than traditional methods (ZNCC + SGM, MC-CNN + SGM) in all the datasets.
    \item The supervised method (OmniMVS) achieves the best performance on OmniThings and OmniHouse, while ours achieve the best performance on Sunny. On the other hand, we argue that comparing our method with OmniMVS is unfair because the supervised method additionally utilizes expensive data labels in the training.
    We demonstrate the performance of OmniMVS for a more straightforward evaluation of our performance.
    \item The Sunny is a small-scale dataset, while OmniThings is a large-scale dataset. We notice that the proposed unsupervised solution achieves better performance in the small-scale dataset.
\end{itemize}

\begin{figure*}[ht]\centering
\includegraphics[scale=0.365]{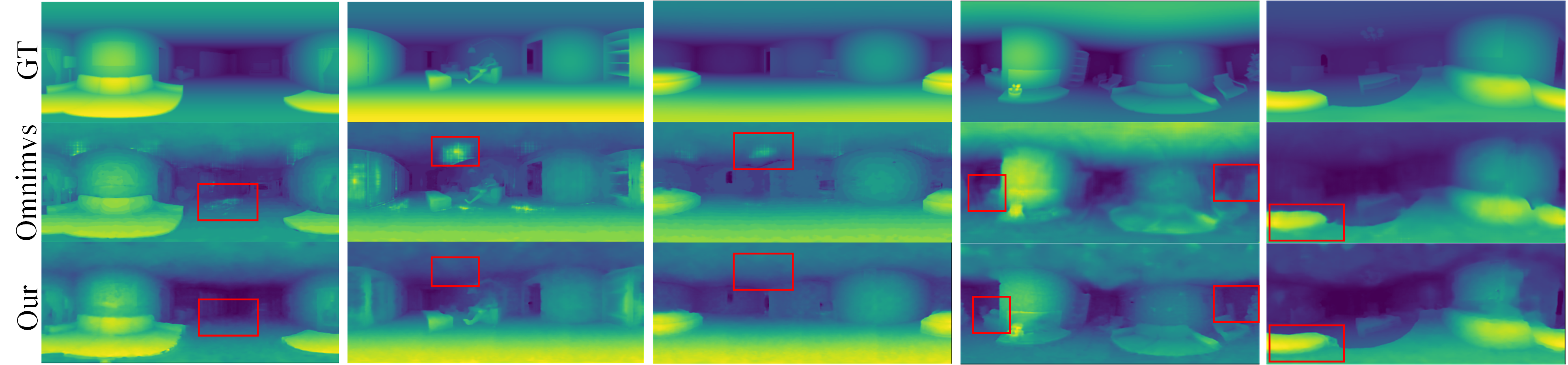}
\caption{ \label{Fig5} Qualitative comparison with the state-of-the-art supervised solution.}
\end{figure*}

Besides, we illustrate the qualitative comparisons in Fig. \ref{Fig5}, where we compare our solution with the supervised solution. In visual appearance, OmniMVS and ours achieve competitive performance, while ours can yield better prediction in the indoor textureless regions as highlighted by the red rectangles. To a certain extent, the supervised method is affected by the size of the training data, which leads to its inability to have a good learning experience for textureless areas, while the unsupervised method can quickly grasp its own geometric matching relationship.
\begin{table}[ht]\centering
\scriptsize
\caption{\footnotesize \centering  Comparison with the state-of-the-arts. The best results are in Bold and the second-best figures are in underlined.}
\label{lable1}
\scalebox{0.9}
{
\begin{tabular}{c|c|c|c|c|c|c|c}
\hline
\textbf{Dataset} & \textbf{Method} & \textbf{Mode}  & \textbf{ \textgreater 1} & \textbf{ \textgreater 3} & \textbf{ \textgreater 5} & \textbf{MAE}  & \textbf{RMS}\\
\hline
            & OmniMVS  &S &  { \textbf{49.3}}   & { \textbf{21.8}}  & {\textbf{13.1}}     & {\textbf{2.9}}   & {\textbf{5.9}}\\
OmniThings  & ZNCC+SGM  &- & 72.6 & 54.0 & 45.6 & 10.5 & 16.5\\
            & MC-CNN+SGM  &- & 67.3 & 47.5 & 39.9  &8.7 &13.7\\
            & \textbf{Un-OmniMVS} &U & \underline{51.7} & \underline{31.8} & \underline{23.3}  &\underline{4.1} &\underline{7.2}\\
\hline  \hline 
            & OmniMVS  &S & { \textbf{29.8}} & { \textbf{7.9}}  & { \textbf{4.3}}  & { \textbf{1.3}} & { \textbf{2.7}}  \\
OmniHouse   & ZNCC+SGM  &-& 44.1 & 20.6 & 13.6 & 3.1 & 7.1\\
            & MC-CNN+SGM &- & 38.0 & 15.9 & 9.5  &2.1 &4.2 \\
            & \textbf{Un-OmniMVS}&U & \underline{34.2} & \underline{11.9} & \underline{7.4}  &\underline{1.8} &\underline{3.7} \\
\hline     \hline     
            & OmniMVS  &S   & 41.5 & 16.6  & 8.7 & 1.9 & 4.6 \\
Sunny       & ZNCC+SGM &- & 52.0 & 21.5 & 11.0 & 2.5 & 5.4\\
            & MC-CNN+SGM  &-& \underline{39.4} & \underline{11.7} & \underline{6.3}  &\underline{1.8} &\underline{4.5}\\
            & \textbf{Un-OmniMVS}&U & { \textbf{35.6}} & { \textbf{11.0}} & { \textbf{6.0}}  & { \textbf{1.6}} & { \textbf{4.0}}\\
\hline 
\end{tabular}
}
\end{table}

\subsubsection{Generalization Experiment}
The supervised methods learn to regress the depth from the supervision of depth labels. On the contrary, the proposed unsupervised solution learns to model the spherical stereo-matching relationship from the unlabeled data, making it work well even in different datasets. To validate this argument, we select the pre-trained models on OmniThings and OmniHouse to verify the generalization capability of the Sunny dataset. As can be seen from Table \ref{lable3}, here, the proposed solution is able to achieve better performance than supervised methods in these cross-dataset evaluations. It enables our model a better transferable capability to different datasets. In fields such as robotics and autonomous driving, superior generalization capabilities are more competitive in the face of changing external environments.
\subsubsection{Ablation Experiment}
We conduct ablation experiments to evaluate the importance of the proposed modules on OmniHouse.
\begin{itemize}
    \item \textbf{Frequency Attention Module:} As illustrated in Table \ref{lable4}, we replace the frequency attention module with a simple skip connection (w/o FAM) and spatial domain attention module (w/ SAM) to prove its effectiveness. The results show 0.5 and 0.2 decreases in RMS when adopting this module (w/ FAM), which could offer better feature perception about the non-local information in the frequency domain. At the same time, we compare our feature extraction with other mainstream methods. We find that these methods consume more time, but do not achieve better performance. To a certain extent, it can explain the inadaptability of downsampling to fisheye images.
    \item \textbf{Variance-Based Cost Volume:} As can be seen from Table \ref{lable5}, the proposed variance-based light cost volume is better than other cost volumes \cite{r17,r21}. In addition to the performance, the results demonstrate our superiority in spatial size and inference speed.
\end{itemize}

\begin{table}[ht]\centering
\footnotesize
\caption{\footnotesize \centering  The generalization difference between the supervised solution and the unsupervised one.}
\label{lable3}
\scalebox{0.85}{
  \begin{tabular}{c|c|c|c|c|c|c}
\hline
\textbf{Pretrained} & \textbf{Method}  & \textbf{ \textgreater 1} & \textbf{ \textgreater 3} & \textbf{ \textgreater 5} & \textbf{MAE}  & \textbf{RMS}\\
\hline
              
OmniThings  & OmniMVS & 51.3 & 23.6 & 16.0 &3.2 &6.6\\ 
            & \textbf{Un-OmniMVS} &  { \textbf{34.1}} & { \textbf{11.1}} &{ \textbf{6.8}}  & { \textbf{1.6}}  & { \textbf{3.7}} \\
\hline  
OmniHouse   & OmniMVS  & 59.6   & 26.5 & 16.4  & 2.9 & 5.3\\
            & \textbf{Un-OmniMVS}  & { \textbf{29.6}}   & { \textbf{8.5}}  &{ \textbf{6.3 }}    & { \textbf{1.4}}   & { \textbf{3.3}}  \\
\hline         

\end{tabular}
}
\end{table}

\begin{table}[ht]\centering
\footnotesize
\caption{\footnotesize \centering  Verify the effectiveness of our frequency domain convolution module.}
\label{lable4}
\scalebox{0.95}{
  \begin{tabular}{c|c|c|c|c|c}
\hline
\textbf{Method}  & \textbf{ \textgreater 1} & \textbf{ \textgreater 3} & \textbf{ \textgreater 5} & \textbf{MAE}  & \textbf{RMS}   \\ 
\hline
Ours w/o FDAM   & 37.9 & 13.5 & 8.5    & 2.2   & 4.2     \\
Ours w/ SDAM   & 38.3 & 13.0 & 7.9    & 2.1   & 3.9     \\
Ours w/ FDAM   & { \textbf{34.2}} & { \textbf{11.9}} & { \textbf{7.4}}  &{ \textbf{1.8}} &{ \textbf{3.7}} \\
\hline 
\hline
\textbf{Method}  & \textbf{ \textgreater 1} & \textbf{ \textgreater 3}  & \textbf{MAE}  & \textbf{RMS} & \textbf{Runtime}  \\ 
\hline
Ours w/ Unet   & 37.1 & 15.8     & 2.2   & 4.3  & 0.76     \\
Ours w/ Resnet   & 39.0 & 14.7  & 2.3   & 4.3   & 0.72    \\
Ours    & { \textbf{34.2}} & { \textbf{11.9}}   &{ \textbf{1.8}} &{ \textbf{3.7}} &{ \textbf{0.33}}\\
\hline 
\end{tabular}
}
\end{table}
\begin{table}[ht]\centering
\footnotesize
\caption{\footnotesize \centering  Verifying the effectiveness and efficiency of our cost volume generation based on variance.}
\label{lable5}
\scalebox{0.95}{
 \begin{tabular}{c|c|c|c|c}
\hline
\textbf{Method}  & \textbf{MAE}  & \textbf{RMS} & \textbf{Size} & \textbf{Runtime} \\ 
\hline
Ours w/ cat \cite{r17}   & 2.3   & 4.4   & $4C \times H \times W$ & 0.44 \\
Ours w/ cat \cite{r21}    & 2.0   & 4.0   & $2C \times H \times W$ & 0.39 \\
Ours w/ Variance    &{ \textbf{1.8}} &{ \textbf{3.7}} &\textbf{$C \times H \times W$} & \textbf{0.33}\\
\hline 
\end{tabular}
}
\end{table}
\begin{figure}
\centering
\includegraphics[scale=0.26]{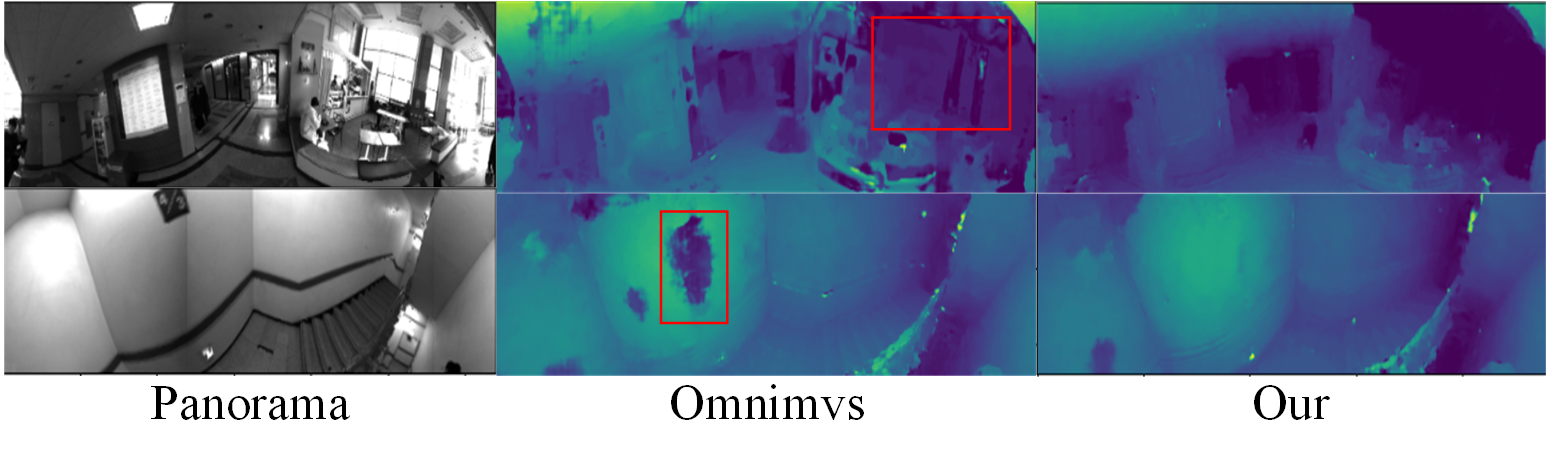}
\caption{\label{realsult}Illustrations of real-world depth estimation. The number of depth samples is set to 32 due to memory limitations.}
\end{figure}
\subsection{Real-World Evaluation}
To demonstrate the effectiveness of our method, we explore the prediction capability of our method and OmniMVS on real-world data \cite{r17}. The camera in the device is installed horizontally at 1.5m and vertically at 2.3m. Since the real data lacks corresponding depth labels, we adopt the pre-trained models for OmniMVS and ours to carry on evaluation. To ensure the fairness of the comparison, we set the same input resolution, output resolution, and the number of depth sampling (We choose D=32 due to memory limitations). It can be seen from Fig. \ref{realsult} that our network can not only achieve better results on synthetic datasets but also produce good prediction results on real data, including some detailed information as highlighted by the red rectangles. However, OmniMVS fails to accomplish effective predictions due to the huge domain gap between the synthetic data and real data, which also suggests that the proposed unsupervised solution has better generalizability by overcoming this gap.
\section{Conclusion}
\label{section5}
We propose the first unsupervised omnidirectional MVS framework by establishing pseudo-stereo supervision. Particularly, we synthesize two panoramas to form a pseudo-stereo pair with a special camera pose. Then the photometric consistency is leveraged to build the unsupervised constraint. 
In addition, we propose Un-OmniMVS with two efficient components: 
a new feature extractor with frequency attention and a variance-based light cost volume. Experimental results show our superiority over other SoTA solutions, especially better generalization in real-world data.

\normalem
\bibliographystyle{plain}
\bibliography{ref}
\end{document}